\documentclass{article}
\usepackage{PRIMEarxiv}

\usepackage{cite}
\usepackage{amsmath,amssymb,amsfonts}
\usepackage{algorithmic}
\usepackage{graphicx}
\usepackage{textcomp}
\usepackage{xcolor}

\usepackage{booktabs}    
\usepackage{tikz}
\usepackage{multicol}
\usepackage{colortbl}
\usepackage{caption}
\usepackage{makecell}
\usepackage{appendix}

\usepackage{caption}
\usepackage[round]{natbib}
\usepackage{booktabs} \newcommand{\ra}[1]{\renewcommand{\arraystretch}{#1}}

\title{Classifying German Language Proficiency Levels Using Large Language Models
}

\author{
  Elias-Leander Ahlers \\
  Computer Science Department\\
  University of Münster \\
  Münster, Germany\\
  \texttt{elias.ahlers@uni-muenster.de} \\
  \And
  Witold Brunsmann \\
  Computer Science Department\\
  University of Münster \\
  Münster, Germany\\
  \texttt{witold.brunsmann@uni-muenster.de}
   \And
  Malte Schilling \\
  Computer Science Department\\
  University of Münster \\
  Münster, Germany\\
  \texttt{malte.schilling@uni-muenster.de} \\
}

\begin{document}

\maketitle

\begin{abstract}
Assessing language proficiency is essential for education, as it enables instruction tailored to learners needs. This paper investigates the use of Large Language Models (LLMs) for automatically classifying German texts according to the Common European Framework of Reference for Languages (CEFR) into different proficiency levels. To support robust training and evaluation, we construct a diverse dataset by combining multiple existing CEFR-annotated corpora with synthetic data. We then evaluate prompt-engineering strategies, fine-tuning of a LLaMA-3-8B-Instruct model and a probing-based approach that utilizes the internal neural state of the LLM for classification. Our results show a consistent performance improvement over prior methods, highlighting the potential of LLMs for reliable and scalable CEFR classification.
\end{abstract}

%\begin{IEEEkeywords}
%large language models, CEFR classification, prompt engineering, fine-tuning, automated language assessment
%\end{IEEEkeywords}
\section{Introduction}
Large Language Models (LLMs) such as GPT, LLaMA, and Gemini have demonstrated impressive capabilities in processing and generating human-like text \citep{Minaee2024}. They are now broadly used, for example, as chatbots, for providing customer support  \citep{xu2024rag} or as tutoring systems \citep{kahl2024evaluating}.

However, their use as specialized classifiers for complex, fine-grained tasks remains underexplored. One such task is the classification of language proficiency levels as, for example, defined by the Common European Framework of Reference for Languages (CEFR). The CEFR framework defines six proficiency levels (A1--C2), each reflecting distinct linguistic capabilities. Accurate classification into these levels requires models to capture subtle differences in vocabulary, grammar, and discourse. Traditional approaches for CEFR classification rely on manually engineered features and classical machine learning methods such as SVMs or neural networks \citep{Caines2020, Szuuegyi2019}. While effective to a certain degree, these methods have shown to be labor-intensive and often limited in scalability.

This paper investigates whether LLMs can serve as effective alternatives for CEFR classification. We approach this through systematically evaluating and comparing different strategies that employ LLMs, using prompt engineering, fine-tuning, and a probing based classifier. Prompt engineering explores how the phrasing of model instructions affects performance. Fine-tuning adapts a pre-trained model to the specific classification task. The probing based approach extracts the internal neural state of the Large Language Model of the last layer and utilizes this as an input for a neural network based classifier which is trained in a supervised fashion on a subset of the dataset. 

A challenge lies in the lack of a comprehensive, balanced dataset for German CEFR classification. Existing corpora often focus on selected levels (e.g., A2--B2) or lack standardized labeling. To address this, we, as a first contribution, develop a new dataset by combining established corpora with synthetic data generation. This allows us to achieve a balanced distribution across the different CEFR levels (A1--C2) which enables a more reliable evaluation of classification strategies. Our experiments show that using LLMs when supported by a carefully constructed dataset and optimized through prompt engineering, fine-tuning and probing offer a powerful and scalable alternative to traditional approaches for CEFR classification and related linguistic assessment tasks. Considering the internal state of the LLM improves the results to a certain degree.
The remainder of the paper is organized as follows. The next section gives an overview on datasets with a focus on German as well as introduces the current state-of-the-art machine learning approaches. The third section explains the construction of a balanced and augmented dataset as well as the different LLM-based classification approaches, we present our results in a comparative evaluation and conclude with a brief discussion.

%Building on these insights, we then investigate fine-tuning approaches, demonstrating how targeted training can further enhance classification performance. Using the LLaMA-3-8B-Instruct model as our foundation, we achieve a weighted F1 score of 0.7686 and a weighted accuracy of 77.3\% across all six CEFR levels, demonstrating significant improvements over traditional classification methods.

\section{Related Work}
In this section, we provide background on key aspects relevant to our study: first, existing datasets for German CEFR classification. Secondly, traditional approaches to modeling language proficiency using classical machine learning methods which we will use as a baseline for comparison.

\subsection{Overview of CEFR Datasets in German}
Research on CEFR classification for German relies on a small number of publicly available, annotated corpora. In this section, we describe the main datasets (overview, see Table \ref{tab:cefr_distribution}) used in our study, focusing on their structure, proficiency level coverage, and suitability for classification tasks.

\subsubsection{Falko Corpus}
The \textit{Falko corpus} \citep{reznicek2010falko}, collected and annotated at Humboldt University Berlin, was designed to facilitate research in corpus linguistics and second language acquisition. It contains a collection of German texts written under controlled conditions by both advanced foreign learners and native speakers, and it includes rich linguistic annotations which include a test score that can be mapped to CEFR levels.

\subsubsection{MERLIN Corpus}
The \textit{MERLIN corpus} \citep{boyd2014merlin} was specifically designed to illustrate CEFR levels using authentic learner texts in German, Italian, and Czech. The corpus covers all six CEFR levels (A1--C2), with a particularly strong representation of the A2, B1, and B2 levels.

The corpus includes a variety of text types, e.g., essays and summaries, collected as part of standardized language tests, ensuring consistency and quality while providing diversity of samples. All texts were annotated with CEFR levels by language experts, providing reliable classifications for our dataset that could be directly included without further processing.

\begin{table}[b!]
\centering
\ra{1.1}
\captionsetup{format=hang}
\caption{Distribution of German CEFR-labeled texts across datasets and levels. Text counts are shown per source and aggregated per CEFR level.}
%\resizebox{0.7\columnwidth}{!}
{
\begin{tabular}{lcccccc}
    \toprule
    \textbf{Source} & \textbf{A1} & \textbf{A2} & \textbf{B1} & \textbf{B2} & \textbf{C1} & \textbf{C2} \\
    \midrule
    Falko EssayL2     &     &     &     & 83  & 84  & 81   \\
    Falko SummaryL1   &     &     &     &     &     & 58   \\
    Falko SummaryL2   &     &     &     &     & 53  & 53   \\
    \midrule
    MERLIN            & 57  & 306 & 331 & 293 & 42  & 4    \\
    \midrule
    Synthetic         & 122 &     &     &     &     &      \\
    \midrule
    \textbf{Our Dataset (1,567)} & \textbf{179} & \textbf{306} & \textbf{331} & \textbf{376} & \textbf{179} & \textbf{196} \\
    \bottomrule
\end{tabular}
}
\label{tab:cefr_distribution}
\end{table}

\subsubsection{Additional Resources}
While CEFR-labeled resources for German are relatively limited in size and scope, there are significantly larger datasets available for English language proficiency. Most notably, the CEFR-SP dataset \citep{Arase2022} contains approximately $17,000$ English sentences labeled by language experts. Other significant English datasets include the Cambridge Learner Corpus and the EF-Cambridge Open Language Database \citep{Geertzen2014}. 

Overall, these datasets demonstrate the potential scale and variety possible in CEFR-level classification resources, though similar comprehensive collections for German remain a work in progress.

\subsection{Traditional Approaches to CEFR Classification}
Existing research on automatic CEFR classification in German has mainly focused on feature-based approaches using classical machine learning method. Two main approaches in this area are by Vajjala and Loo \citep{Vajjala2018}, extended by Caines et al. \citep{Caines2020}, and, secondly, by Szüügyi et al. \citep{Szuuegyi2019}.

Vajjala and Rama \citep{Vajjala2018} proposed to combine handcrafted linguistic features with neural network classifiers to classify German CEFR levels. The authors extracted a variety of features from the texts, including lexical, syntactic, and semantic indicators, which they used to train a neural network classifier. This work, has been extended by Caines et al. \citep{Caines2020}, who incorporated additional linguistic features and applied it to two other languages (Spanish and English) as well. Their system achieved a weighted F1 score of $0.702$ for German, demonstrating the effectiveness of feature-based methods in CEFR classification.

Szüügyi et al. \citep{Szuuegyi2019} used a Support Vector Machine (SVM) to classify German CEFR levels on linguistic features from the texts, which again included lexical, syntactic, and semantic features. Furthermore, they tested incorporating a Multi-Layer Perceptron (MLP) classifier. In their work, they simplified the problem by merging the CEFR levels only into the three broader categories: A(A1,A2), B(B1,B2), and C(C1,C2). For these three classes, they report an accuracy of $82\%$.

These methods rely heavily on features that are hand-crafted by experts and partially operate only on reduced versions of the CEFR scale. This limits their applicability and scalability to real-world applications.

\section{Methods}
We evaluate several leading Large Language Models for the CEFR classification task, taking into account factors such as model size, inference speed, and performance on German texts (see \ref{subsec:performance_comparison}). After comprehensive testing and an evaluation of different prompts, we selected the LLaMA-3-8B-Instruct model \citep{LLaMA3} as our base model for fine-tuning.
In the following sections, we will introduce the main parts of our approach: First, the construction of a dedicated and balanced CEFR dataset for German which requires data augmentation. Second, the development and evaluation of various prompt engineering strategies. Third, we present a fine-tuning pipeline based on LLaMA-3-8B-Instruct to assess the effect of supervised adaptation on classification performance. Fourth, we introduce a probing-based analysis in which a neural classifier is trained on the final-layer activations of the LLaMA-3 model to investigate how CEFR-relevant linguistic features are encoded in its internal representations. 

\subsection{Construction of a Balanced CEFR Dataset}
 \label{sec:dataset}
The creation of a comprehensive German CEFR classification dataset presented several challenges due to the limited public availability of CEFR-labeled texts and in particular, as existing datasets tend to be sparse and imbalanced. We addressed this by combining multiple existing corpora along with synthetic generated data to ensure a balanced representation across proficiency levels as required for a robust evaluation.

As a first source, we used the already introduced Falko corpus as it contains high-quality texts for CEFR classification. We selected three subcorpora to ensure a diverse range of text types and proficiency levels:
\begin{itemize}
    \item \textbf{EssayL2}, consists of argumentative essays written by advanced foreign learners of German (CEFR levels between B2 to C2). Topics include feminism, financial policy, and crime. All texts were written under controlled conditions within a 90-minute time limit.
    \item \textbf{SummaryL1}, contains academic summaries from native speakers, which we used as reference samples for the C2 level.
    \item \textbf{SummaryL2}, includes similar summaries written by learners of German as a second language.
\end{itemize}
Together, these subcorpora offer valuable examples of high-proficient language use and were instrumental in anchoring the upper end of the CEFR scale in our dataset. The EssayL2 subcorpus contained texts from B2 to C2, each annotated with metadata including the author's C-test score, an integer ranging from 0 to 100 measuring German proficiency. We used the Falko authors' mapping scheme to assign CEFR levels to each text based on these scores (see Table \ref{tab:c_test_mapping}).

\begin{table}[htb]
\centering

\caption{Mapping of FALKO C-Test Scores to CEFR Levels}
%\resizebox{0.8\columnwidth}{!}
{
\begin{tabular}{p{3cm}c}
    \toprule
    \textbf{Score Range} & \textbf{CEFR Level} \\
    \midrule
    60--79 & B2 \\
    80--89 & C1 \\
    90--100 & C2 \\
    \bottomrule
\end{tabular}
}
\label{tab:c_test_mapping}
\end{table}

We also added the MERLIN corpus which provides a diverse set of texts across all CEFR levels, with a strong focus on A2, B1, and B2 levels. The corpus includes essays and summaries written by learners of German as a second language and is annotated with CEFR levels by language experts. 

To address the remaining gap at the lower end of the proficiency spectrum, in particular the lack of A1-level texts, we employed a synthetic data generation approach using a large language model. We chose the Claude 3.5 Sonnet model \citep{claude3.5sonnet}, an advanced LLM developed by Anthropic. While we used LLaMA-3-8B-Instruct as our classification model, we opted for Claude for data generation for several reasons. First, Claude has demonstrated superior ability in following complex instructions and maintaining consistent quality \citep{Huang2024}. Additionally it shows good performance on German language understanding \citep{park2024}. Second, as a larger model with a broader training data coverage, it allows for more nuanced and authentic generation of German learner texts \citep{kim2024}. The prompt engineering process for data generation underwent several iterations until a desired output quality was reached. Initial attempts produced texts that were too simplistic or thematically biased toward primary school topics. To overcome this, we refined the approach and adjusted the prompts to avoid stereotypical beginner subjects, incorporated explicit CEFR A1 criteria from the Goethe Institute, and used prompts written in German, which produced noticeably improved results. The final prompt is included in the appendix (appendix section~\ref{appendix:generation}). Using this approach, we generated 122 synthetic texts. Each sample was manually reviewed to ensure appropriateness and quality before inclusion in the dataset.

The final dataset consists of 1,567 texts spanning across all six CEFR levels (A1-C2). Table~\ref{tab:cefr_distribution} shows the distribution of texts by source and level.

\subsection{Prompt Engineering for CEFR Classification}
We developed and refined a prompting strategy through several iterations, each aimed at improving the model's classification accuracy. Prompt engineering plays a crucial role in aligning the model's behavior with the task objective by structuring inputs and instructions in a way that leverages the model's internal representations \citep{white2023promptengineering}. A well-designed prompt should guide the model to focus on relevant textual features in order to produce consistent label predictions, even without access to explicit training signals. In particular, prompts should clearly define the task as well as the expected output format to ensure reliable and interpretable model responses.

We started with an \textbf{English base prompt} (see Appendix \ref{appendix:prompts} for full prompts, Table \ref{tab:prompt_variants}) that formulates the classification task in explicit terms: it instructs the model to assess a given text and assign a proficiency level according to the CEFR scale. The prompt specifies both the decision criterion (CEFR classification) and the required response format (a single level label, such as A1--C2), ensuring that the model focuses on the intended task and avoids generating additional explanations. 

Next, we switched to German language prompts as Pires et al. found \citep{Pires2019} that LLMs often show a better performance when the prompt language matches the task language. In initial tests, the model frequently included justifications for its classification when using the German prompt. This might skew the results and, therefore, we appended an instruction that the model should suppress any additional explanation or reasoning for the classification. The final German prompt (\textbf{German zero-shot prompt}, given in Appendix \ref{appendix:prompts}, Table \ref{tab:prompt_variants}) emphasizes both the classification standard and the required output format, helping to reduce noise in the model's predictions.

Building on this, we implemented a few-shot prompting strategy (\textbf{German few-shot prompt}, see Table \ref{tab:prompt_variants} in Appendix \ref{appendix:prompts}). The prompt included the base instruction and incorporated an example for each CEFR level, demonstrating typical text characteristics at each level. Examples were carefully selected from previously misclassified texts to highlight common challenges and improve the model´s ability to distinguish between adjacent levels. By explicitly showing example texts at each level, the few-shot format provided additional context for the LLM and served as an implicit decision guide.

\begin{table}[tb]
\centering
\ra{1.1}
\captionsetup{format=hang}
\caption{Hyperparameters used for fine-tuning the LLaMA-3-8B-Instruct model.}
%\resizebox{0.9\columnwidth}{!}
{
\begin{tabular}{@{}ll@{}}
\toprule
\textbf{Hyperparameter} & \textbf{Value} \\
\midrule
\textbf{Training Settings} & \\
Learning rate & 2e-4 \\
Number of epochs & 5 (cutoff after 3) \\
Batch size & 1 \\
Gradient accumulation steps & 1 \\
Warmup steps & 400 \\
Weight decay & 0.001 \\
Learning rate scheduler & Linear \\
\addlinespace
\textbf{LoRA Configuration} & \\
LoRA rank ($r$) & 32 \\
LoRA alpha & 32 \\
LoRA dropout & 0.03 \\
LoRA target modules & \makecell[l]{q\_proj, k\_proj, v\_proj, o\_proj,\\gate\_proj, up\_proj, down\_proj} \\
Bias & none \\
Gradient checkpointing & True \\
Use DoRA / RSLoRA & False \\
\bottomrule
\end{tabular}
}
\label{tab:finetuning_hyperparams}
\end{table}

\subsection{Fine-Tuning LLaMA-3 for CEFR Classification}
To complement our prompt-based experiments, we fine-tuned a Large Language Model specifically on the CEFR classification task. Fine-tuning allows a model to adapt more closely to domain-specific data and structure \citep{devlin2019bert}. It can yield significant improvements in performance over prompt-based strategies. 

We selected the LLaMA-3-8B-Instruct model \citep{LLaMA3} as the foundation for fine-tuning. This choice was driven by several factors: the model has a balanced size of 8 billion parameters, it is openly available, and has demonstrated strong capabilities in instruction following which is essential for adapting to our structured classification task. Moreover, the model has shown promising performance in German-language understanding.

Fine-tuning follows a supervised learning approach, for which we used the dataset introduced in Section~\ref{sec:dataset}. The dataset was split into a training (154 samples per CEFR level) and a test set (25 samples per level), ensuring balanced representation across proficiency levels. Training samples were formatted following the LLaMA-3 input structure, with system prompts, user messages, and assistant responses clearly separated by special tokens (using the LLaMA-3 tokenizer and paired with their corresponding CEFR labels). For fine-tuning, we used LoRA \citep{Hu2021}.
Training was performed on a NVIDIA RTX A6000 GPU, with the process completing in approximately 20 minutes. Fine-tuning was conducted using the hyperparameters given in Table \ref{tab:finetuning_hyperparams}.

%\begin{table}[htb]
%\centering
%\ra{1.1}
%\captionsetup{format=hang}
%\caption{Hyperparameters used for fine-tuning the LLaMA-3-8B-Instruct model.}
%\resizebox{0.6\columnwidth}{!}{
%\begin{tabular}{@{}ll@{}}
%\toprule
%\textbf{Hyperparameter} & \textbf{Value} \\
%\midrule
%Learning rate & 2e-4 \\
%Number of epochs & 3 \\
%Batch size & 1 \\
%Gradient accumulation steps & 1 \\
%Warmup steps & 400 \\
%Weight decay & 0.001 \\
%Learning rate scheduler & Linear \\
%\bottomrule
%\end{tabular}}
%\label{tab:finetuning_hyperparams}
%\end{table}

\subsection{Probing-based Neural Classifier} 
Large language models (LLMs) contain internal states which refers to representations that typically remain hidden during inference. Probing of internal states allows to trace the internal processing of the LLM and has shown to contain additional information \citep{lindsey2025biology,ameisen2025circuit}. As one example, internal states have previously been applied in hallucination detection, where classifiers were trained using the internal representations as an input and were able to identify hallucinated outputs with higher accuracy than response-based methods \citep{azaria2023internal, ridder2024hallurag}. This demonstrates that internal states contain rich information about the model's internal reasoning, beyond what is expressed in its generated text. 

For all experiments, we used LLaMA-3-8B. We chose the base version rather than the Instruct model, since our setup does not require text generation but relies solely on extracting contextualized embedding vectors (CEVs) for the given text samples. The instruction-tuned variant is optimized for producing conversational and helpful continuations, which could introduce unwanted biases in the internal representations. To avoid such effects, we used the base model for all CEV extractions. Input to the LLM were the tokenized texts from the dataset. These sequences of texts were passed to the model for processing. Internal states of the last layer for the last token of a text  were extracted using the Transformers library \citep{huggingface2025}, which provides direct access to the final-layer embeddings for all tokens in a sequence.

We trained a multi-layer perceptron (MLP) classifier on the CEFR-annotated dataset described in a supervised fashion. % mapping the CEV activations towards the CEFR class labels. 
The final MLP architecture consists of four fully connected layers (\texttt{input\_size}-1024-512-256-6) with ReLU activations and softmax in the output layer, following previous probing architectures \citep{ridder2024, su2024unsupervised}. Training used a learning rate of $1\times10^{-3}$ and an L2 regularization of $0.001$ (hyperparameters were tuned using a grid search over architecture, learning rate, and regularization). The network receives the contextualized embedding vectors (CEVs) from the last hidden layer of the LLaMA-3-8B model as input, with the assigned CEFR proficiency label as the target output.  These embeddings served as input features for the classifier. As the data sets are small, we used five fold cross validation to evaluate the classification performance.

%$ (comparable to the $20%$ used in MIND). We employed early stopping, terminating after a maximum of 800 epochs or when the validation loss failed to improve for 30 consecutive epochs. The checkpoint with the lowest validation loss was retained for testing.

\begin{figure*}[b!]
    \centering
    \begin{minipage}{0.32\textwidth}
    \centering
    \textbf{(a) English Base Prompt}\\
    \vspace{2px}
    \resizebox{0.95\textwidth}{!}{
    \begin{tabular}{c|cccccc}
    & \multicolumn{6}{c}{Predicted} \\
    Actual & A1 & A2 & B1 & B2 & C1 & C2 \\
    \hline
    A1 & \cellcolor[rgb]{0.9,1,0.9}3 & \cellcolor[rgb]{0.97,1,0.97}1 & \cellcolor[rgb]{0.2,0.8,0.2}21 & 0 & 0 & 0 \\
    A2 & 0 & 0 & \cellcolor[rgb]{0.2,0.8,0.2}25 & 0 & 0 & 0 \\
    B1 & 0 & 0 & \cellcolor[rgb]{0.22,0.81,0.22}24 & \cellcolor[rgb]{0.97,1,0.97}1 & 0 & 0 \\
    B2 & 0 & 0 & \cellcolor[rgb]{0.46,0.88,0.46}17 & \cellcolor[rgb]{0.68,0.93,0.68}8 & 0 & 0 \\
    C1 & 0 & 0 & \cellcolor[rgb]{0.72,0.95,0.72}7 & \cellcolor[rgb]{0.4,0.86,0.4}18 & 0 & 0 \\
    C2 & 0 & 0 & 0 & \cellcolor[rgb]{0.2,0.8,0.2}25 & 0 & 0 \\
    \end{tabular}}
    \end{minipage}
    \begin{minipage}{0.32\textwidth}
    \centering
    \textbf{(b) German Zero-Shot Prompt}\\
    \vspace{2px}
    \resizebox{0.95\textwidth}{!}{
    \begin{tabular}{c|cccccc}
    & \multicolumn{6}{c}{Predicted} \\
    Actual & A1 & A2 & B1 & B2 & C1 & C2 \\
    \hline
    A1 & \cellcolor[rgb]{0.8,0.96,0.8}5 & \cellcolor[rgb]{0.72,0.95,0.72}7 & \cellcolor[rgb]{0.76,0.95,0.76}6 & 0 & 0 & 0 \\
    A2 & 0 & \cellcolor[rgb]{0.84,0.97,0.84}4 & \cellcolor[rgb]{0.5,0.89,0.5}14 & 0 & 0 & 0 \\
    B1 & 0 & \cellcolor[rgb]{0.97,1,0.97}1 & \cellcolor[rgb]{0.23,0.81,0.23}23 & \cellcolor[rgb]{0.97,1,0.97}1 & 0 & 0 \\
    B2 & 0 & 0 & \cellcolor[rgb]{0.28,0.83,0.28}21 & \cellcolor[rgb]{0.84,0.97,0.84}4 & 0 & 0 \\
    C1 & 0 & 0 & \cellcolor[rgb]{0.76,0.95,0.76}6 & \cellcolor[rgb]{0.34,0.85,0.34}19 & 0 & 0 \\
    C2 & 0 & 0 & \cellcolor[rgb]{0.76,0.95,0.76}6 & \cellcolor[rgb]{0.34,0.85,0.34}19 & 0 & 0 \\
    \end{tabular}}
    \end{minipage}
    \begin{minipage}{0.32\textwidth}
    \centering
    \textbf{(c) German Few-Shot Prompt}\\
    \vspace{2px}
    \resizebox{0.95\textwidth}{!}{
    \begin{tabular}{c|cccccc}
    & \multicolumn{6}{c}{Predicted} \\
    Actual & A1 & A2 & B1 & B2 & C1 & C2 \\
    \hline
    A1 & \cellcolor[rgb]{0.46,0.88,0.46}15 & \cellcolor[rgb]{0.8,0.96,0.8}5 & \cellcolor[rgb]{0.8,0.96,0.8}5 & 0 & 0 & 0 \\
    A2 & \cellcolor[rgb]{0.9,0.98,0.9}3 & \cellcolor[rgb]{0.66,0.93,0.66}9 & \cellcolor[rgb]{0.54,0.9,0.54}13 & 0 & 0 & 0 \\
    B1 & 0 & \cellcolor[rgb]{0.97,1,0.97}1 & \cellcolor[rgb]{0.42,0.87,0.42}16 & \cellcolor[rgb]{0.72,0.95,0.72}7 & \cellcolor[rgb]{0.97,1,0.97}1 & 0 \\
    B2 & 0 & 0 & 0 & \cellcolor[rgb]{0.28,0.83,0.28}21 & \cellcolor[rgb]{0.93,0.99,0.93}2 & \cellcolor[rgb]{0.93,0.99,0.93}2 \\
    C1 & 0 & 0 & 0 & \cellcolor[rgb]{0.69,0.94,0.69}8 & \cellcolor[rgb]{0.76,0.95,0.76}6 & \cellcolor[rgb]{0.58,0.91,0.58}11 \\
    C2 & 0 & 0 & 0 & \cellcolor[rgb]{0.97,1,0.97}1 & \cellcolor[rgb]{0.93,0.99,0.93}2 & \cellcolor[rgb]{0.26,0.82,0.26}22 \\
    \end{tabular}}
    \end{minipage}
    %\caption{Confusion matrices for different prompt engineering approaches using the LLaMA-3-8B-Instruct model: (a) English Base Prompt, (b) German Zero-Shot Prompt, (c) German Few-Shot Prompt. Each matrix visualizes predicted CEFR levels (columns) against true labels (rows), with cell shading indicating prediction density.}
    \caption{Confusion matrices for different prompt engineering approaches using the LLaMA-3-8B-Instruct model: (a) English Base Prompt, (b) German Zero-Shot Prompt, (c) German Few-Shot Prompt. Each matrix visualizes predicted CEFR levels (columns) against true labels (rows), with cell shading indicating prediction density. The mean classification distances are: English Base Prompt = $1.120$, German Zero-Shot Prompt = $1.051$, German Few-Shot Prompt = $0.467$.}
    \label{fig:confusion_prompts}
\end{figure*}

\section{Results}
In this section, we present the results of our CEFR classification experiments, structured into four main parts: prompt engineering, comparison of different LLMs, probing approach, and fine-tuning. We evaluated the model's performance using standard classification metrics including accuracy, precision, recall, and F1 score. Additionally, we introduced a \textit{group accuracy} metric to account for the continuous nature of CEFR levels, considering classifications of adjacent levels as correct. As another metric, we defined a \textit{mean classification distance}, which quantifies the average gap between predicted and true language proficiency levels. Distances are assigned as integer penalties, with adjacent levels counting as $1$ and more distant errors receiving proportionally larger penalties. In contrast to binary accuracy, this approach provides more nuanced evaluation by weighting the severity of classification errors. This approach better reflects the practical application of CEFR classification, as boundaries between levels are at least to a certain degree subjective and fluid. For selected setups, we additionally report confusion matrices to highlight common misclassifications between adjacent CEFR levels. Together, these results offer insights into the strengths and limitations of LLM-based CEFR classification and help identify promising directions for further improvement.

\subsection{Prompt Engineering Results}
To evaluate the impact of prompt design on classification accuracy, we tested the LLaMA-3-8B-Instruct model using several prompt variants applied to our CEFR-labeled dataset (see Section~\ref{sec:dataset}), which includes a balanced distribution of $1,567$ learner texts across all six CEFR levels. These included an English base prompt, a German zero-shot prompt, and a German few-shot prompt. Table~\ref{tab:performance_summary} summarizes the performance of each prompt type.

\begin{table}[tb]
    \centering
    \captionsetup{format=hang}
    \caption{Performance comparison across different prompt types, the neural network-based probing classifier, and the fine-tuning approach (group accuracy includes adjacent levels).}
    \vspace{5px}
    %\resizebox{\columnwidth}{!}
    {
        \begin{tabular}{p{4.5cm}cc}
            \toprule
            \textbf{Prompt Name} & \textbf{Accuracy} & \textbf{Group Accuracy} \\
            \midrule
            English Base Prompt & 23.3\% & 64.6\% \\
            German Zero-Shot Prompt & 33.3\% & 75.3\% \\
            German Few-Shot Prompt & 59.3\% & 94.0\% \\
            \midrule
            Probing NN Classifier & 65,83\% & 99.2\% \\
            \midrule
            Fine-tuned LLaMA-3-8B-Instruct & 76.7\% & 100.0\% \\
            \bottomrule
        \end{tabular}
    }
    \label{tab:performance_summary}
\end{table}

The \textbf{English Base Prompt} served as our initial method for instructing the model to classify texts according to CEFR levels. While straightforward in its instruction to classify texts according to CEFR levels, performance was limited (overall performance, see Table \ref{tab:performance_summary}, for detailed metrics on each level see appendix \ref{appendix:metrics}, Table \ref{tab:metrics_all}). This held also true for specific levels, for instance, the model achieved a precision of 0.471 and recall of 0.640 for intermediate levels like B1. The corresponding confusion matrix revealed a tendency to default to middle-range levels (B1--B2), suggesting that the model struggled to differentiate between extreme proficiency levels (see Fig. \ref{fig:confusion_prompts} (a)). Particularly notable is the model's tendency to misclassify A1 and A2 texts as a B1 level, with $21$ out of $25$ A1 texts and all $25$ A2 texts being incorrectly classified as B1. This showed as well in the \textit{mean classification distance} of $1.12$. The systematic error pattern suggested that the English-language prompt might be limiting the model's ability to accurately assess German language proficiency.

The transition to a German prompt marked a significant improvement in classification performance. The \textbf{German Zero-Shot Prompt} demonstrated more nuanced classification abilities, particularly for intermediate proficiency levels. This approach showed improved differentiation between adjacent levels, with the confusion matrix revealing more balanced distribution of classifications across CEFR levels (see Fig. \ref{fig:confusion_prompts} (b)). Overall, these improvement suggests that language alignment between the prompt and the classification task plays a crucial role in model performance (\textit{mean classification distance} improved as well to $1.051$).

%The German prompt's performance revealed an interesting pattern in the handling of extreme proficiency levels. For A1 texts, the model achieved a precision of $0.833$ and recall of $0.600$, significantly outperforming the English prompt (detailed metrics are given in the appendix, table \ref{tab:metrics_zero_shot}). 

Next, we observed further substantial improvements in classification accuracy across all proficiency levels for the \textbf{German Few-Shot Prompt}. The confusion matrix revealed a more diagonal pattern, indicating better discrimination between adjacent proficiency levels (see Fig. \ref{fig:confusion_prompts} (c); detailed metrics are provided in appendix \ref{appendix:metrics}, Table \ref{tab:metrics_all})). Especially interesting was the model's improved handling of extreme proficiency levels. A1 classification showed a precision of $0.833$ and recall of $0.600$, while C2 classification achieved a precision of $0.629$ and recall of $0.880$ (\textit{mean classification distance} of $0.467$ highlights this improvement as well). This balanced performance across the proficiency spectrum suggests that the few-shot examples helped the model develop a more nuanced understanding of the characteristics associated with each CEFR level. 

\begin{figure*}[b!]
    \centering
    \begin{tikzpicture}[scale=0.85]
        % Define colors
        \definecolor{barcolor1}{RGB}{31,119,180}
        \definecolor{barcolor2}{RGB}{255,127,14}

        % Draw axes
        \draw[->] (0,0) -- (15,0) node[right] {Models};
        \draw[->] (0,0) -- (0,5.5);
        \node[anchor=south] at (0,5.8) {Performance};

        % Y-axis labels (scaled to 50%)
        \foreach \y in {0,10,20,30,40,50}
            \draw (0,\y/10) node[left] {\y} -- (-0.1,\y/10);

        % Combined Data (12 models)
        \foreach [count=\i] \model/\accuracy/\groupaccuracy in {
            {Mistral-7B-Instruct}/21.33/58.00,
            {mistral-nemo}/28.00/66.70,
            {qwen-2.5-7b-instruct}/28.00/68.70,
            {Mixtral-8x22B-Instruct-v0.1}/32.00/65.33,
            {llama-3.1-8b-instruct}/34.70/77.30,
            {google/gemma-2-9b-it}/36.00/74.67,
            {google/gemma-2-27b-it}/38.00/77.33,
            {Qwen/Qwen2-7B-Instruct}/38.00/80.67,
            {Meta-LLaMA-3-70B-Instruct}/44.67/92.67,
            {Qwen/Qwen2-72B-Instruct}/54.67/98.00,
            {\underline{Meta-Llama-3-8B}}/59.33/94.00
        } {
            % Draw bars (scaled by half)
            \fill[barcolor1] (1.3*\i-1.0,0) rectangle (1.3*\i-0.6,\accuracy/20);
            \fill[barcolor2] (1.3*\i-0.5,0) rectangle (1.3*\i-0.1,\groupaccuracy/20);

            % Labels
            \node[below, text width=2.5cm, align=center, rotate=45, anchor=north east] at (1.3*\i-0.1,0) {\footnotesize\model};
            \node[above right, xshift=1pt] at (1.3*\i-1.4,\accuracy/20) {\footnotesize\accuracy\%};
            \node[above right, xshift=1pt] at (1.3*\i-0.9,\groupaccuracy/20) {\footnotesize\groupaccuracy\%};
        }

        % Legend
        \node[anchor=north west, inner sep=0, outer sep=0] at (2,6.2) {
            \begin{tikzpicture}
                \fill[barcolor1] (0,0) rectangle (0.3,0.3);
                \node[right] at (0.4,0.15) {Accuracy (\%)};
                \fill[barcolor2] (3,0) rectangle (3.3,0.3);
                \node[right] at (3.4,0.15) {Group Accuracy (\%)};
            \end{tikzpicture}
        };
    \end{tikzpicture}
    \caption{Performance comparison of different language models on CEFR classification, showing both exact accuracy and group accuracy (includes adjacent levels), sorted by Accuracy (names are model names as found on huggingface).}
    \label{fig:llm-performance-comparison}
\end{figure*}
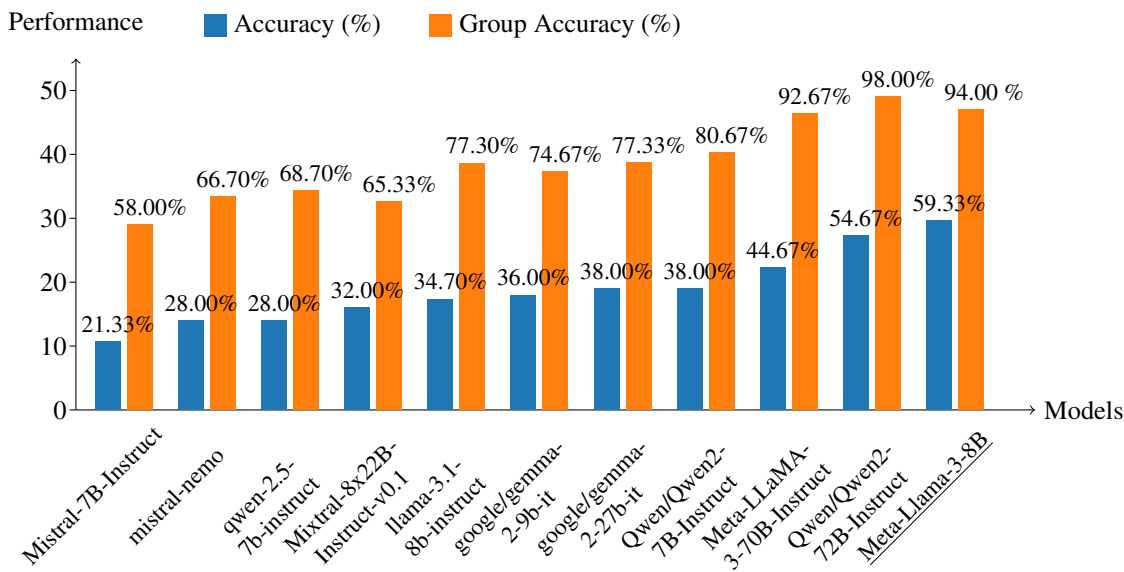

\subsection{Performance Comparison Across LLMs}
\label{subsec:performance_comparison}
To evaluate the baseline capabilities of various pre-trained language models on the CEFR classification task, we tested twelve widely-used instruction-tuned LLMs on our balanced dataset using the German few-shot prompt. Models were selected to represent different architectures, parameter sizes, and training sources. Each model was prompted in German using the same CEFR classification instruction to ensure consistency across evaluations. We recorded both exact classification accuracy and a form of a more relaxed group-level accuracy (that includes adjacent levels).
Figure~\ref{fig:llm-performance-comparison} illustrates the performance comparison across all evaluated models.

\begin{table}[b!]
    \centering
    \captionsetup{format=hang}
    \caption{Performance metrics for the neural network based classifier across CEFR levels. The classifier was trained on the internal CEV states of the LLM as an input and produces the corresponding class labels. Results are mean results on the test set for $5$-fold cross validation.}
    \vspace{5px}
    %\resizebox{\columnwidth}{!}
    {
        \begin{tabular}{p{2.5cm}ccc}
            \toprule
            \textbf{Class} & \textbf{Precision} & \textbf{Recall} & \textbf{F1 Score} \\
            \midrule
            A1 & 0.889 & 0.800 & 0.842 \\
            A2 & 0.600 & 0.750 & 0.667 \\
            B1 & 0.524 & 0.550 & 0.537 \\
            B2 & 0.632 & 0.600 & 0.615 \\
            C1 & 0.647 & 0.550 & 0.595 \\
            C2 & 0.700 & 0.700 & 0.700 \\
            \midrule
            \textbf{Weighted Avg} & \textbf{0.665} & \textbf{0.658} & \textbf{0.659} \\
            \bottomrule
        \end{tabular}
    }
    \label{tab:probe-metrics}
\end{table}

\begin{figure}[b!]
    \centering
    \resizebox{0.4\columnwidth}{!}{
    \begin{tabular}{c|cccccc}
    & \multicolumn{6}{c}{Predicted} \\
    Actual & A1 & A2 & B1 & B2 & C1 & C2 \\
    \hline
    A1 & \cellcolor[rgb]{0.35,0.85,0.35}16 & \cellcolor[rgb]{0.82,0.97,0.82}4 & \cellcolor[rgb]{1,1,1}0 & \cellcolor[rgb]{1,1,1}0 & \cellcolor[rgb]{1,1,1}0 & \cellcolor[rgb]{1,1,1}0 \\
    A2 & \cellcolor[rgb]{0.86,0.97,0.86}2 & \cellcolor[rgb]{0.42,0.87,0.42}15 & \cellcolor[rgb]{0.80,0.96,0.80}3 & \cellcolor[rgb]{1,1,1}0 & \cellcolor[rgb]{1,1,1}0 & \cellcolor[rgb]{1,1,1}0 \\
    B1 & \cellcolor[rgb]{1,1,1}0 & \cellcolor[rgb]{0.88,0.98,0.88}6 & \cellcolor[rgb]{0.55,0.90,0.55}11 & \cellcolor[rgb]{0.84,0.97,0.84}3 & \cellcolor[rgb]{1,1,1}0 & \cellcolor[rgb]{1,1,1}0 \\
    B2 & \cellcolor[rgb]{1,1,1}0 & \cellcolor[rgb]{1,1,1}0 & \cellcolor[rgb]{0.75,0.95,0.75}7 & \cellcolor[rgb]{0.45,0.88,0.45}12 & \cellcolor[rgb]{1,1,1}0 & \cellcolor[rgb]{0.85,0.97,0.85}1 \\
    C1 & \cellcolor[rgb]{1,1,1}0 & \cellcolor[rgb]{1,1,1}0 & \cellcolor[rgb]{1,1,1}0 & \cellcolor[rgb]{0.84,0.97,0.84}4 & \cellcolor[rgb]{0.50,0.89,0.50}11 & \cellcolor[rgb]{0.80,0.96,0.80}5 \\
    C2 & \cellcolor[rgb]{1,1,1}0 & \cellcolor[rgb]{1,1,1}0 & \cellcolor[rgb]{1,1,1}0 & \cellcolor[rgb]{1,1,1}0 & \cellcolor[rgb]{0.70,0.94,0.70}6 & \cellcolor[rgb]{0.40,0.87,0.40}14 \\
    \end{tabular}
    }
    \caption{Confusion matrix for the neural network based classifier, highlighting a reduced confusion between adjacent CEFR levels.}
    \label{fig:confusion_probe}
\end{figure}

\subsection{Performance of the Probing Neural Network Classifier}
A neural network based classifier was trained that uses the full internal state of the last layer as an input and maps this to a class label. Test results for the classifier are shown in Table \ref{tab:probe-metrics} when using five-fold cross validation. The accuracy achieved was $65.83\%$ with a precision of $0.665$,  recall of $0.658$, , and an F1 score of  $0.659$. Additionally, the group accuracy was $99.2\%$, indicating that almost all misclassifications occurred between adjacent levels only. The corresponding confusion matrix is given in Figure \ref{fig:confusion_probe} which further shows that there are only small deviations. Overall, the neural network classifier shows an improvement across all metrics when compared to querying the LLM using only language (see Table \ref{tab:performance_summary} for comparison of the results). This indicates that the internal states contain additional supporting information.

\subsection{Fine-tuned Model Performance Analysis}
To evaluate the benefits of model adaptation, we fine-tuned the LLaMA-3-8B-Instruct model on our balanced CEFR dataset which showed the highest accuracy (Fig. \ref{fig:llm-performance-comparison}). This aims for the model to learn domain-specific patterns in the selected language and better distinguish between subtle linguistic features associated with each CEFR level. The fine-tuning process revealed consistent and stable improvement in model performance. The training loss curve showed a steady decrease across epochs, with significant improvements in the early stages followed by gradual stabilization. The stabilization of training loss around epoch 3 suggested optimal convergence without overfitting, validating our choice of training parameters. This pattern was particularly encouraging given the relatively small size of our training dataset, indicating very fast but still effective learning from limited examples.

\begin{figure}[tb]
    \centering
    \resizebox{0.4\columnwidth}{!}{
    \begin{tabular}{c|cccccc}
    & \multicolumn{6}{c}{Predicted} \\
    Actual & A1 & A2 & B1 & B2 & C1 & C2 \\
    \hline
    A1 & \cellcolor[rgb]{0.28,0.83,0.28}21 & \cellcolor[rgb]{0.84,0.97,0.84}4 & \cellcolor[rgb]{1,1,1}0 & \cellcolor[rgb]{1,1,1}0 & \cellcolor[rgb]{1,1,1}0 & \cellcolor[rgb]{1,1,1}0 \\
    A2 & \cellcolor[rgb]{0.9,0.98,0.9}3 & \cellcolor[rgb]{0.37,0.86,0.37}18 & \cellcolor[rgb]{0.84,0.97,0.84}4 & \cellcolor[rgb]{1,1,1}0 & \cellcolor[rgb]{1,1,1}0 & \cellcolor[rgb]{1,1,1}0 \\
    B1 & \cellcolor[rgb]{1,1,1}0 & \cellcolor[rgb]{0.93,0.99,0.93}2 & \cellcolor[rgb]{0.28,0.83,0.28}21 & \cellcolor[rgb]{0.93,0.99,0.93}2 & \cellcolor[rgb]{1,1,1}0 & \cellcolor[rgb]{1,1,1}0 \\
    B2 & \cellcolor[rgb]{1,1,1}0 & \cellcolor[rgb]{1,1,1}0 & \cellcolor[rgb]{0.84,0.97,0.84}4 & \cellcolor[rgb]{0.42,0.87,0.42}16 & \cellcolor[rgb]{0.8,0.96,0.8}5 & \cellcolor[rgb]{1,1,1}0 \\
    C1 & \cellcolor[rgb]{1,1,1}0 & \cellcolor[rgb]{1,1,1}0 & \cellcolor[rgb]{1,1,1}0 & \cellcolor[rgb]{0.84,0.97,0.84}4 & \cellcolor[rgb]{0.28,0.83,0.28}21 & \cellcolor[rgb]{1,1,1}0 \\
    C2 & \cellcolor[rgb]{1,1,1}0 & \cellcolor[rgb]{1,1,1}0 & \cellcolor[rgb]{1,1,1}0 & \cellcolor[rgb]{1,1,1}0 & \cellcolor[rgb]{0.72,0.95,0.72}7 & \cellcolor[rgb]{0.37,0.86,0.37}18 \\
    \end{tabular}
    }
    \caption{Confusion matrix for the fine-tuned LLaMA-3-8B model, showing improved accuracy and reduced confusion between neighboring CEFR levels.}
    \label{fig:confusion_finetuned}
\end{figure}

The fine-tuned LLaMA-3-8B-Instruct model demonstrated substantial improvements across all six proficiency levels, achieving a weighted F1 score of 0.769 across the complete CEFR spectrum. The resulting confusion matrix, shown in Figure~\ref{fig:confusion_finetuned}, demonstrates substantially improved classification accuracy and a more balanced performance across all CEFR levels. The distribution of misclassifications shows a clear pattern of only adjacent-level errors. This pattern is particularly clear in the B1-C1 range, where all misclassifications occur between neighboring proficiency levels. For instance, B2 texts were only misclassified to either B1 or C1, suggesting that the model maintains a reasonable understanding of the proficiency spectrum even when making errors.
This pattern of adjacent-level errors is particularly relevant when considering the continuous nature of language proficiency assessment. When accounting for this through our group accuracy metric, which considers classifications of adjacent levels as partially correct, the model achieves a perfect group accuracy of $100\%$ on the test set. This result indicates that even when the model makes errors, it maintains a fundamentally good understanding of the proficiency spectrum. An improvided \textit{mean classification distance} of as low as $0.233$ for the fine-tuned model further demonstrates that adaptation worked quite well.

\begin{table}[b!]
    \centering
    \captionsetup{format=hang}
    \caption{Performance metrics of the fine-tuned model across CEFR levels.}
    \vspace{5px}
    %\resizebox{\columnwidth}{!}
    {
        \begin{tabular}{p{2.5cm}ccc}
            \toprule
            \textbf{Class} & \textbf{Precision} & \textbf{Recall} & \textbf{F1 Score} \\
            \midrule
            A1 & 0.875 & 0.840 & 0.857 \\
            A2 & 0.750 & 0.720 & 0.735 \\
            B1 & 0.724 & 0.840 & 0.778 \\
            B2 & 0.727 & 0.640 & 0.681 \\
            C1 & 0.636 & 0.840 & 0.724 \\
            C2 & 1.000 & 0.720 & 0.837 \\
            \midrule
            \textbf{Weighted Avg} & \textbf{0.785} & \textbf{0.767} & \textbf{0.769} \\
            \bottomrule
        \end{tabular}
    }
    \label{tab:finetune_metrics}
\end{table}

The fine-tuned model showed distinct performance patterns across different proficiency levels, revealing both strengths and areas for potential improvement. At the extremes of the proficiency spectrum, the model showed particularly strong performance, with A1 and C2 levels achieving the highest F1 scores of 0.857 and 0.837 respectively (Table \ref{tab:finetune_metrics}).

 The A1 level performance is particularly noteworthy, combining high precision (0.875) with strong recall (0.840). This balanced performance suggests robust capabilities in identifying beginner-level texts, a crucial ability for practical applications in language assessment. The perfect precision (1.000) achieved at the C2 level, although with lower recall (0.720), indicates high reliability when the model identifies advanced-level texts. In the intermediate range, we observed more nuanced performance patterns. B1 level texts were classified with strong consistency, achieving an F1 score of 0.778 with notably high recall (0.840). This suggests effective identification of intermediate proficiency markers. However, the B2 level presented the most significant challenge, with an F1 score of 0.681, reflecting the inherent difficulty of distinguishing this transitional proficiency level.

% Moved discussion to external file. TODO Include in Arxiv version

\section{Discussion and Conclusion}
The presented approach demonstrates the effectiveness of Large Language Models (LLMs) in classifying German texts according to CEFR proficiency levels. Through careful prompt engineering and fine-tuning of the LLaMA-3-8B-Instruct model, we achieved a weighted F1 score of $0.769$ for the fine-tuned model which represents a substantial improvement over previous state-of-the-art results, such as the $0.702$ reported by Caines \& Buttery~\citep{Caines2020}. Notably, our model achieved perfect \textit{group accuracy} ($100\%$), demonstrating excellent performance in assigning texts to the correct general CEFR region—that is, either the exact level or a directly adjacent one. This represents a significant improvement over prior work using broader level groupings. For instance, Szüügyi et al. \citep{Szuuegyi2019} reported $82\%$ accuracy using a three-level classification (A, B, C) based on a Linear SVM and manual feature engineering. When applying the evaluation scheme to our results (see Table~\ref{fig:confusion_finetuned}), our fine-tuned model achieves $76.7\%$ accuracy and $100\%$ group accuracy, indicating a notable improvement. Such performance appears particularly valuable in practical applications, where the boundaries between CEFR levels often have to be considered fluid and subjective. The fine-tuned model's ability to consistently place texts within the correct region of the proficiency spectrum, even if not always at the exact level, demonstrates its reliability as a tool for language assessment support.

A neural network based classifier that was trained on the internal states of the LLaMA3 model showed an improvement compared to pure prompting approaches when applied to the same LLM. The internal state of the model appears to contain further valuable information that support language level classification. Importantly, the fine-tuning approach outperformed the neural network probing classifier which demonstrates that without fine-tuning the LLaMA model does not contain sufficient information for a better performance.

Overall, our results demonstrate several key advances: strong performance in distinguishing extreme proficiency levels (A1 and C2), consistent handling of intermediate levels, and reliable classification of adjacent CEFR levels. These results suggest significant potential for LLMs to support language assessment processes, particularly when combined with human expertise. 
 
As language assessment tools continue to evolve, the integration of LLM-based approaches shows promise in enhancing the efficiency and reliability of the CEFR classification process. Finally, a promising direction lies in leveraging LLMs not only for classification but also for generation of texts on a specific CEFR level or rewriting existing ones to match a target proficiency level. One challenge for this is the limited availability of high-quality, labeled training data, as generating or annotating such texts typically requires the involvement of language assessment experts. The presented classification model could support such approaches as it can serve as a reliable tool for automatically labeling or validating large volumes of generated data. This can facilitate the creation of synthetic datasets that are both diverse and consistently aligned with CEFR standards.

\bibliography{References}

\appendices
\section{Prompt for Generation of Synthetic Data}\label{appendix:generation}
German prompt used for synthetic data generation, following definition by \citep{goetheCefr}:

\begin{quotation}
\textit{Bitte generiere Texte mit dem CEFR Niveau A1. Diese sollten länger (Circa 600 Wörter) sein. Versuche Themen zu finden, welche nicht mit Schule/Kindheit in Verbindung stehen. Es sollte sich um Texte für Deutsch Sprachler mit dem Level A1 handeln. \\
        Hier ist eine Definition des A1 Levels: \\
        Kann vertraute, alltägliche Ausdrücke und ganz einfache Sätze verstehen und verwenden, die auf die Befriedigung konkreter Bedürfnisse zielen.
        Kann sich und andere vorstellen und anderen Leuten Fragen zu ihrer Person stellen - z. B. wo sie wohnen, welche Leute sie kennen oder welche Dinge sie haben - und kann auf Fragen dieser Art Antwort geben.
        Kann sich auf einfache Art verständigen, wenn die Gesprächspartner langsam und deutlich sprechen und bereit sind zu helfen.}
    \end{quotation}
    \label{appendix:synthetic-prompt}

\section{Learning curves for Fine-Tuning the LLM}\label{appendix:training}

\begin{figure}[tbh]
    \centering
    \begin{tikzpicture}[scale=0.7]
        \definecolor{traincolor}{RGB}{0,114,178}
        \definecolor{valcolor}{RGB}{230,159,0}
        \definecolor{acccolor}{RGB}{213,94,0}
        
        \draw[->] (0,0) -- (10.5,0) node[right] {Epoch};
        \draw[->] (0,0) -- (0,6) node[above] {Loss};
        \draw[->] (10.5,0) -- (10.5,6) node[above] {Accuracy (\%)};
        
        \foreach \x in {0,1,2,3,4,5}
            \draw (\x*2,0) node[below] {\x} -- (\x*2,-0.1);
        
        \foreach \y in {0,0.5,1,1.5,2,2.5}
            \draw (0,\y*2) node[left] {\y} -- (-0.1,\y*2);
        
        \foreach \y in {20,40,60,80,100}
            \draw (10.5,\y/100*2.5*2) node[right] {\y} -- (10.5+0.1,\y/100*2.5*2);
        
        \draw[gray, dashed, thin] (3*2,0) -- (3*2,6);
        
        \draw[traincolor, smooth, thick] plot coordinates {
            (0.042*2,  2.0360*1.88)
            (0.125*2,  1.9521*1.88)
            (0.208*2,  1.9062*1.88)
            (0.292*2,  1.8674*1.88)
            (0.375*2,  1.7220*1.88)
            (0.458*2,  1.6876*1.88)
            (0.542*2,  1.6640*1.88)
            (0.625*2,  1.6432*1.88)
            (0.708*2,  1.6499*1.88)
            (0.792*2,  1.6210*1.88)
            (0.875*2,  1.5949*1.88)
            (0.958*2,  1.5546*1.88)
            (1.042*2,  1.5299*1.88)
            (1.125*2,  1.4816*1.88)
            (1.208*2,  1.4437*1.88)
            (1.292*2,  1.4006*1.88)
            (1.375*2,  1.3930*1.88)
            (1.458*2,  1.3792*1.88)
            (1.542*2,  1.3926*1.88)
            (1.625*2,  1.3695*1.88)
            (1.708*2,  1.3985*1.88)
            (1.792*2,  1.3470*1.88)
            (1.875*2,  1.3043*1.88)
            (1.958*2,  1.2305*1.88)
            (2.042*2,  1.1778*1.88)
            (2.125*2,  1.0953*1.88)
            (2.208*2,  1.0505*1.88)
            (2.292*2,  0.9750*1.88)
            (2.375*2,  0.9559*1.88)
            (2.458*2,  0.9558*1.88)
            (2.542*2,  0.9676*1.88)
            (2.625*2,  0.9840*1.88)
            (2.708*2,  0.9964*1.88)
            (2.792*2,  0.9590*1.88)
            (2.875*2,  0.9118*1.88)
            (2.958*2,  0.8589*1.88)
            (3.042*2,  0.8075*1.88)
            (3.125*2,  0.7490*1.88)
            (3.208*2,  0.6600*1.88)
            (3.292*2,  0.5929*1.88)
            (3.375*2,  0.5787*1.88)
            (3.458*2,  0.5741*1.88)
            (3.542*2,  0.5758*1.88)
            (3.625*2,  0.5621*1.88)
            (3.708*2,  0.5652*1.88)
            (3.792*2,  0.5592*1.88)
            (3.875*2,  0.5346*1.88)
            (3.958*2,  0.4935*1.88)
            (4.042*2,  0.4380*1.88)
            (4.125*2,  0.4255*1.88)
            (4.208*2,  0.3831*1.88)
            (4.292*2,  0.3447*1.88)
            (4.375*2,  0.3302*1.88)
            (4.458*2,  0.3514*1.88)
            (4.542*2,  0.3527*1.88)
            (4.625*2,  0.3632*1.88)
            (4.708*2,  0.3494*1.88)
            (4.792*2,  0.3560*1.88)
            (4.875*2,  0.3614*1.88)
            (4.958*2,  0.3678*1.88)
        };
        
        \draw[valcolor, thick, dashed, mark=o, mark options={scale=0.6, solid}] plot coordinates {
            (0.050*2, 1.9473*1.88)
            (0.150*2, 1.8589*1.88)
            (0.250*2, 1.7706*1.88)
            (0.350*2, 1.6512*1.88)
            (0.450*2, 1.5006*1.88)
            (0.550*2, 1.3499*1.88)
            (0.650*2, 1.2605*1.88)
            (0.750*2, 1.2339*1.88)
            (0.850*2, 1.2073*1.88)
            (0.950*2, 1.1685*1.88)
            (1.050*2, 1.1170*1.88)
            (1.150*2, 1.0656*1.88)
            (1.250*2, 1.0465*1.88)
            (1.350*2, 1.0614*1.88)
            (1.450*2, 1.0763*1.88)
            (1.550*2, 1.0871*1.88)
            (1.650*2, 1.0938*1.88)
            (1.750*2, 1.1005*1.88)
            (1.850*2, 1.0880*1.88)
            (1.950*2, 1.0547*1.88)
            (2.050*2, 1.0214*1.88)
            (2.150*2, 1.0019*1.88)
            (2.250*2, 0.9972*1.88)
            (2.350*2, 0.9925*1.88)
            (2.450*2, 0.9902*1.88)
            (2.550*2, 0.9903*1.88)
            (2.650*2, 0.9904*1.88)
            (2.750*2, 0.9833*1.88)
            (2.850*2, 0.9681*1.88)
            (2.950*2, 0.9529*1.88)
            (3.050*2, 0.9617*1.88)
            (3.150*2, 0.9977*1.88)
            (3.250*2, 1.0336*1.88)
            (3.350*2, 1.0411*1.88)
            (3.450*2, 1.0159*1.88)
            (3.550*2, 0.9908*1.88)
            (3.650*2, 0.9665*1.88)
            (3.750*2, 0.9431*1.88)
            (3.850*2, 0.9198*1.88)
            (3.950*2, 0.9037*1.88)
            (4.050*2, 0.8963*1.88)
            (4.150*2, 0.8888*1.88)
            (4.250*2, 0.8813*1.88)
            (4.350*2, 0.8738*1.88)
            (4.450*2, 0.8663*1.88)
            (4.550*2, 0.8588*1.88)
            (4.650*2, 0.8513*1.88)
            (4.750*2, 0.8438*1.88)
            (4.850*2, 0.8363*1.88)
            (4.950*2, 0.8288*1.88)
        };
        
        \draw[acccolor, smooth, thick, dotted] plot coordinates {
            (0.050*2, 0.7251*1.88)
            (0.150*2, 0.7418*1.88)
            (0.250*2, 0.7584*1.88)
            (0.350*2, 0.8655*1.88)
            (0.450*2, 1.0629*1.88)
            (0.550*2, 1.2603*1.88)
            (0.650*2, 1.3436*1.88)
            (0.750*2, 1.3129*1.88)
            (0.850*2, 1.2821*1.88)
            (0.950*2, 1.3223*1.88)
            (1.050*2, 1.4334*1.88)
            (1.150*2, 1.5445*1.88)
            (1.250*2, 1.6139*1.88)
            (1.350*2, 1.6416*1.88)
            (1.450*2, 1.6694*1.88)
            (1.550*2, 1.6416*1.88)
            (1.650*2, 1.5582*1.88)
            (1.750*2, 1.4749*1.88)
            (1.850*2, 1.4944*1.88)
            (1.950*2, 1.6166*1.88)
            (2.050*2, 1.7389*1.88)
            (2.150*2, 1.7945*1.88)
            (2.250*2, 1.7834*1.88)
            (2.350*2, 1.7723*1.88)
            (2.450*2, 1.7612*1.88)
            (2.550*2, 1.7500*1.88)
            (2.650*2, 1.7388*1.88)
            (2.750*2, 1.7749*1.88)
            (2.850*2, 1.8582*1.88)
            (2.950*2, 1.9416*1.88)
            (3.050*2, 1.9777*1.88)
            (3.150*2, 1.9666*1.88)
            (3.250*2, 1.9555*1.88)
            (3.350*2, 1.9333*1.88)
            (3.450*2, 1.9000*1.88)
            (3.550*2, 1.8667*1.88)
            (3.650*2, 1.8750*1.88)
            (3.750*2, 1.9250*1.88)
            (3.850*2, 1.9750*1.88)
            (3.950*2, 1.9972*1.88)
            (4.050*2, 1.9916*1.88)
            (4.150*2, 1.9860*1.88)
            (4.250*2, 1.9944*1.88)
            (4.350*2, 2.0166*1.88)
            (4.450*2, 2.0389*1.88)
            (4.550*2, 2.0445*1.88)
            (4.650*2, 2.0334*1.88)
            (4.750*2, 2.0223*1.88)
            (4.850*2, 2.0292*1.88)
            (4.950*2, 2.0543*1.88)
        };
        
        \draw[traincolor, thick] (1.5,5.5) -- (2,5.5) node[right,black] {Training Loss};
        \draw[valcolor, thick, dashed] (1.5,5.0) -- (2,5.0) node[right,black] {Validation Loss};
        \draw[acccolor, thick, dotted] (1.5,4.5) -- (2,4.5) node[right,black] {Text Accuracy};
        
    \end{tikzpicture}
    \caption{Training loss (blue), validation loss (orange), and text accuracy (red) during training of the model. The vertical line marks the training cutoff used.}
    \label{fig:class-epoch-loss}
\end{figure}
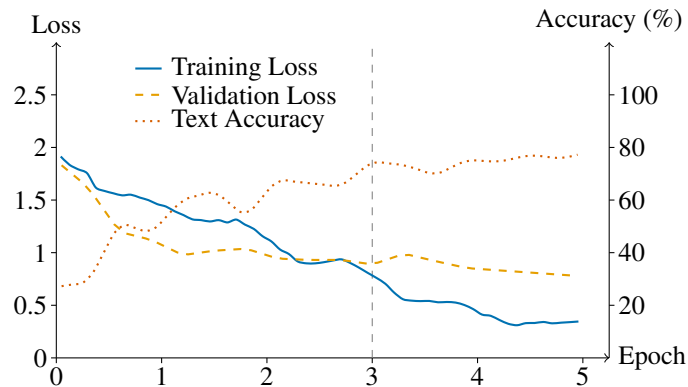

\onecolumn

\section{Detailed Performance Metrics for Prompt Engineering Approaches}\label{appendix:metrics}

\begin{table}[tbh]
\captionsetup{format=hang}
\caption{Performance metrics for different prompts (LLaMA-3-8B-Instruct).}
\label{tab:metrics_all}
\centering
\begin{minipage}[t]{0.32\textwidth}
    \centering
    \caption*{English Base Prompt}
    \resizebox{\linewidth}{!}{%
        \begin{tabular}{p{2.5cm}ccc}
            \toprule
            \textbf{Class} & \textbf{P} & \textbf{R} & \textbf{F1} \\
            \midrule
            A1 & 1.000 & 0.120 & 0.214 \\
            A2 & 0.000 & 0.000 & 0.000 \\
            B1 & 0.255 & 0.960 & 0.403 \\
            B2 & 0.154 & 0.320 & 0.208 \\
            C1 & 0.000 & 0.000 & 0.000 \\
            C2 & 0.000 & 0.000 & 0.000 \\
            \midrule
            \textbf{Avg} & \textbf{0.247} & \textbf{0.467} & \textbf{0.286} \\
            \bottomrule
        \end{tabular}%
    }
\end{minipage}\hfill
\begin{minipage}[t]{0.32\textwidth}
    \centering
    \caption*{German Zero-Shot Prompt}
    \resizebox{\linewidth}{!}{%
        \begin{tabular}{p{2.5cm}ccc}
            \toprule
            \textbf{Class} & \textbf{P} & \textbf{R} & \textbf{F1} \\
            \midrule
            A1 & 1.000 & 0.480 & 0.649 \\
            A2 & 0.579 & 0.440 & 0.500 \\
            B1 & 0.303 & 0.920 & 0.455 \\
            B2 & 0.093 & 0.160 & 0.118 \\
            C1 & 0.000 & 0.000 & 0.000 \\
            C2 & 0.000 & 0.000 & 0.000 \\
            \midrule
            \textbf{Avg} & \textbf{0.327} & \textbf{0.500} & \textbf{0.406} \\
            \bottomrule
        \end{tabular}%
    }
\end{minipage}\hfill
\begin{minipage}[t]{0.32\textwidth}
    \centering
    \caption*{German Few-Shot Prompt}
    \resizebox{\linewidth}{!}{%
        \begin{tabular}{p{2.5cm}ccc}
            \toprule
            \textbf{Class} & \textbf{P} & \textbf{R} & \textbf{F1} \\
            \midrule
            A1 & 0.833 & 0.600 & 0.698 \\
            A2 & 0.600 & 0.360 & 0.450 \\
            B1 & 0.471 & 0.640 & 0.542 \\
            B2 & 0.568 & 0.840 & 0.677 \\
            C1 & 0.546 & 0.240 & 0.333 \\
            C2 & 0.629 & 0.880 & 0.733 \\
            \midrule
            \textbf{Avg} & \textbf{0.609} & \textbf{0.600} & \textbf{0.586} \\
            \bottomrule
        \end{tabular}%
    }
\end{minipage}
\end{table}

\section{Prompt Variations for CEFR Classification}\label{appendix:prompts}
\begin{table*}[h]
\centering
\ra{1.1}
\captionsetup{format=hang}
\caption{Overview of prompt variants used for CEFR Classification. Each prompt was tested in isolation to evaluate its impact on model performance and behavior on our data set.}
\resizebox{\textwidth}{!}{
\begin{tabular}{@{}p{3.3cm}p{5cm}p{5cm}cc@{}}
\toprule
\textbf{Prompt Name} & \textbf{Original Prompt} & \textbf{English Translation} & \textbf{Accuracy} & \textbf{Group Accuracy} \\
\midrule
\textbf{English Base Prompt} & Classify the language level of a given text according to the Common European Framework of Reference for Languages (CEFR). Respond with only the corresponding CEFR level (A1, A2, B1, B2, C1 or C2). & Classify the language level of a given text according to the Common European Framework of Reference for Languages (CEFR). Respond with only the corresponding CEFR level (A1, A2, B1, B2, C1 or C2). & 23.3\% & 64.6\% \\
\midrule
\textbf{German Zero-Shot Prompt} & Bewerte die Sprachkenntnisse des bereitgestellten deutschen Textes gemäß dem Gemeinsamen Europäischen Referenzrahmen für Sprachen (GER/CEFR). Antworte NUR mit der entsprechenden Stufe: A1, A2, B1, B2, C1 oder C2, *keiner* Begründung. & Assess the language proficiency of the given German text according to the CEFR. Respond ONLY with the corresponding level: A1-C2, no justification. & 33.3\% & 75.3\% \\
\midrule
\textbf{German Few-Shot Prompt} & Klassifiziere die Sprachkenntnisse des bereitgestellten deutschen Textes gemäß dem Gemeinsamen Europäischen Referenzrahmen für Sprachen (GER/CEFR). Antworte NUR mit der entsprechenden Stufe: A1, A2, B1, B2, C1 oder C2, NICHT MEHR. Gebe auch *keine* Begründung! Hier sind jeweils Beispiele: A1: [...] A2: [...] B1: [...] B2: [...] C1: [...] C2: [...] & Classify the language proficiency of the following German text according to the CEFR. Respond ONLY with one of the following levels: A1-C2, NOTHING MORE. Do not give any justification! Here are examples for each level: A1: [...] A2: [...] B1: [...] B2: [...] C1: [...] C2: [...] & 59.3\% & 94.0\% \\
\bottomrule
\end{tabular}}
\label{tab:prompt_variants}
\end{table*}

\end{document}